\documentclass[twoside]{article}

\usepackage{epsfig}
\usepackage{amsmath,amssymb}
\usepackage{graphics}

\newcommand{\BEAS}{\begin{eqnarray*}}
\newcommand{\EEAS}{\end{eqnarray*}}
\newcommand{\BEA}{\begin{eqnarray}}
\newcommand{\EEA}{\end{eqnarray}}
\newcommand{\BEQ}{\begin{equation}}
\newcommand{\EEQ}{\end{equation}}
\newcommand{\BIT}{\begin{itemize}}
\newcommand{\EIT}{\end{itemize}}
\newcommand{\BNUM}{\begin{enumerate}}
\newcommand{\ENUM}{\end{enumerate}}
\newcommand{\BA}{\begin{array}}
\newcommand{\EA}{\end{array}}

\newtheorem{proposition}{Proposition}
\newtheorem{lemma}{Lemma}

\newcommand{\tr}{\mathop{ \rm tr}}

\newcommand{\rb}{\mathbb{R}}

\newcommand{\BlackBox}{\rule{1.5ex}{1.5ex}}  % end of proof
\newenvironment{proof}{\par\noindent{\bf Proof\ }}{\hfill\BlackBox\\[2mm]}

\newcommand {\br}[1]{\left(#1\right)}
\newcommand {\sqb}[1]{\left[#1\right]}
\newcommand {\cbr}[1]{\left\{#1 \right\}}
\newcommand {\nm}[1]{\Arrowvert\, #1 \,\Arrowvert}

\newcommand {\abs}[1]{\left\vert\, #1 \,\right\vert}

\newcommand {\ovr}[1]{\frac{1}{#1}}
\newcommand {\Sum}[2]{\sum_{#1=1}^{#2}}

\newcommand{\RR}{\mathbb{R}}    %reals
\newcommand{\Gcal}{\mathcal{G}}
\newcommand{\Kcal}{\mathcal{K}}
\newcommand{\Xcal}{\mathcal{X}} %set of possible inputs x
\newcommand{\Ycal}{\mathcal{Y}} %set of possible outputs y

\newcommand{\xb}{\mathbf{x}}    %training patterns
\newcommand{\yb}{\mathbf{y}}    %training patterns

\newcommand{\ko}{k_{\otimes}}   % tensor product kernel
\newcommand{\Ho}{\mathcal{H}_{\otimes}}
\newcommand{\idm}{I}

\title{Low-rank matrix factorization with attributes}

\author{
Jacob Abernethy\\
Computer Science Division\\
University of California \\
 Berkeley, CA, USA \\
\texttt{jake@eecs.berkeley.edu} \\
\and
Francis Bach\\
Center for Mathematical Morphology\\
Ecole des Mines de Paris\\
Fontainebleau, France\\
\texttt{francis.bach@mines.org} \\[.74cm]
\and
Theodoros Evgeniou\\
INSEAD \\ Fontainebleau, France\\
\texttt{theodoros.evgeniou@insead.edu} \\
\and
Jean-Philippe Vert\\
Center for Computational Biology\\
Ecole des Mines de Paris\\
Fontainebleau, France\\
\texttt{Jean-Philippe.Vert@ensmp.fr} \\[.75cm]
}

\date{Technical report N-24/06/MM\\[.2cm]
Ecole des mines de Paris, France \\[.2cm]
September 2006}

\begin{document}

\maketitle

\begin{abstract}
We develop a new collaborative filtering (CF) method that combines
both previously known users' preferences, i.e. standard CF, as
well as product/user attributes, i.e. classical function
approximation, to predict a given user's interest in a particular
product. Our method is a generalized low rank matrix completion
problem, where we learn a function whose inputs are pairs of
vectors -- the standard low rank matrix completion problem being a
special case where the inputs to the function are the row and
column indices of the matrix. We solve this generalized matrix
completion problem using tensor product kernels for which we also
formally generalize standard kernel properties. Benchmark
experiments on movie ratings show the advantages of our
generalized matrix completion method over the standard matrix
completion one with no information about movies or people, as well
as over standard multi-task or single task learning methods.

\end{abstract}

\section{Introduction}

Collaborative Filtering (CF) refers to the task of predicting
preferences of a given user based on their previously known
preferences as well as the preferences of other users. In a book
recommender system, for example, one would like to suggest new
books to a customer based on what he and others have recently read
or purchased. This can be formulated as the problem of filling a
matrix with customers as rows, objects (e.g., books) as columns,
and missing entries corresponding to preferences that one would
like to infer. In the simplest case, a preference could be a
binary variable (thumbs up/down), or perhaps even a more
quantitative assessment (scale of 1 to 5).

Standard CF assumes that nothing is known about the users or the
objects apart from the preferences expressed so far. In such a
setting the most common assumption is that preferences can be
decomposed into a small number of factors, both for users and
objects, resulting in the search for a low-rank matrix which
approximates the partially observed matrix of preferences. This
problem is usually a difficult non-convex problem for which only
heuristic algorithms exist~\cite{Srebro2003Weighted}.
Alternatively convex formulations have been obtained by relaxing
the rank constraint by constraining the trace norm of the
matrix~\cite{Srebro2005Maximum}.

In many practical applications of CF, however, a description of
the users and/or the objects through attributes (e.g., gender,
age) or measures of similarity is available. In that case it is
tempting to take advantage of both known preferences and
descriptions to model the preferences of users. An important
benefit of such a framework over pure CF is that it potentially
allows the prediction of preferences for new users and/or new
objects. Seen as learning a preference function from examples,
this problem can be solved by virtually any algorithm for
supervised classification or regression taking as input a pair
(user, object). If we suppose for example that a positive definite
kernel between pairs can be deduced from the description of the
users and object, then learning algorithms like support vector
machines or kernel ridge regression can be applied. These
algorithms minimize an empirical risk over a ball of the
reproducing kernel Hilbert space (RKHS) defined by the pairwise
kernel.

Both the rank constraint and the RKHS norm restriction act as
regularization based on prior hypothesis about the nature of the
preferences to be inferred. The rank constraint is based on the
hypothesis that preferences can be modelled by a limited number or
factors to describe users and objects. The RKHS norm constraint
assumes that preferences vary smoothly between similar users and
similar objects, where the similarity is assessed in terms of the
kernel for pairs.

The main contribution of this work is to propose a framework which
combines both regularizations on the one hand, and which
interpolates between the pure CF approach and the pure
attribute-based approaches on the other hand. In particular, the
framework encompasses low-rank matrix factorization for
collaborative filtering, multi-task learning, and classical
regression/classification over product spaces. We show on a
benchmark experiment of movie recommendations that the resulting
algorithm can lead to significant improvements over other
state-of-the-art methods.

\section{Kernels and tensor product spaces}
In this section, we review the classical theory of tensor product
reproducing kernel Hilbert spaces, providing a natural
generalization of finite-dimensional matrices for functions of two
variables. The general setup is as follows. We consider the
general problem of estimating a function $f:\Xcal\times\Ycal
\rightarrow \RR$ given a finite set of observations in $
\Xcal\times\Ycal\times\RR$. We assume that both spaces $\Xcal$ and
$\Ycal$ are endowed with positive semi-definite kernels,
respectively $k:\Xcal\times\Xcal\rightarrow\RR$ and
$g:\Ycal\times\Ycal\rightarrow\RR$, and denote by $\Kcal$ and
$\Gcal$ the corresponding RKHS. A typical application of this
setting is where $\xb\in\Xcal$ is a person, $\yb\in\Ycal$ is a
movie, kernels $k$ and $g$ represent similarities between persons
and movies, respectively, and $f\br{\xb,\yb}$ represents a
person's $\xb$ rating of a movie $\yb$. We note that if $\Xcal$
and $\Ycal$ are finite sets, then $f$ is simply a matrix of size
$\abs{\Xcal}\times\abs{\Ycal}$.

\subsection{Tensor product kernels and RKHS}
We denote by
$\ko:\br{\Xcal\times\Ycal}\times\br{\Xcal\times\Ycal}\rightarrow\RR$
the tensor product kernel, known to be a positive definite kernel \cite[p.70]{Berg1984Harmonic}:
\begin{equation}\label{eq:tensorkernel}
\ko\br{\br{\xb_{1},\yb_{1}},\br{\xb_{2},\yb_{2}}} =
k\br{\xb_{1},\xb_{2}}g\br{\yb_{1},\yb_{2}}\;,
\end{equation}
and by $\Ho$ the associated RKHS. A classical result of Aronszajn~\cite{Aronszajn1950Theory}
states that $\Ho$ is the \emph{tensor product} of the two spaces
$\Kcal$ and $\Gcal$ (denoted $\Kcal\otimes\Gcal$), i.e., $\Ho$ is the completion of all
functions $f: \Xcal\times\Ycal \to \rb$, which can be finitely
decomposed as $f(x,y) = \sum_{k=1}^p u_k(x) v_k(y)$, where $u_k
\in \Kcal$ and $v_k \in \Gcal$, $k=1,\dots, p$.
An \emph{atomic term}  defined as $f(x,y) = u(x)v(y)$, with $u\in\Kcal$ and $v\in\Gcal$,  is usually denoted
$f = u \otimes v$. The space $\Ho$ is  equipped with a norm
such that $\| u \otimes v \|_\otimes = \|u \|_\Kcal \times \| v\|_\Gcal$,
and thus $\| \sum_k u_k \otimes v_k \|^2 = \sum_{k,l} \langle u_k, u_l \rangle_\Kcal
 \langle v_k, v_l \rangle_\Gcal$.

\subsection{Rank}
An element of $\Ho$ can always be decomposed as a possibly infinite sum of
atomic terms of the form $u(x)v(y)$ where $u \in \Kcal$ and $v \in
\Gcal$. We define the \emph{rank} $\text{rank}(f) \in \mathbb{N} \cup
\{\infty\}$ of an element $f$ of $\Ho$ as the minimal number of
atomic terms in any decomposition of $f$, i.e,
$\text{rank}(f)$ is the smallest integer $p$ such that $f$ can be
expanded as:
$$
f\br{\xb,\yb} = \Sum{1}{p}u_{i}\br{\xb}v_{i}\br{\yb}\;,
$$ for some functions $u_{1},\ldots,u_{p} \in\Kcal$ and
$v_{1},\ldots,v_{p}\in\Gcal$, if such an integer $p$ does exist (otherwise, the rank is infinite).

When the two RKHS are  spaces of linear functions on an Euclidean
space, then the tensor product can be identified to the space of
bilinear forms on the product of the two Euclidean spaces, and the
notion of rank coincides with the usual notion of rank for
matrices. We note that an alternative characterization of
$\text{rank}(f)$ is the supremum of the ranks of the matrices $M$
defined by $M_{i,j}=f\br{\xb_{i},\yb_{j}}$ over the choices of
finite sets $\xb_{1},\ldots,\xb_{m}\in\Xcal$ and
$\yb_{1},\ldots,\yb_{p}\in\Ycal$ (see technical annex for a
proof).

\subsection{Trace norm}
Given a rectangular matrix $M$, the rank is not an easy function to optimize or constrain, since it is neither
convex nor continuous. Following the 1-norm approximation to the 0-norm, the trace norm has emerged has
an efficient convex approximation of the rank~\cite{Boyd2001Rank,Srebro2005Maximum}. The trace norm $\|M\|_\ast$ is defined as
the sum of the singular values. This definition is not easy to extend to functional tensor product spaces because it involves
eigen-decompositions. Rather, we use the equivalent formulation
$$\| M \|_\ast = \inf_{ M = U V } \frac{1}{2}(  \|U \|_F^2 + \|V\|_F^2)$$
where $\| U \|_F^2 = \tr UU^\top$ is the squared Frobenius norm.

We thus extend the notion of trace norm  as
$$\|f\|_\ast = \inf_{ f = \sum_{k=1}^\infty u_k \otimes v_k}
\frac{1}{2}\sum_{k=1}^\infty ( \|u_k \|_\Kcal^2 + \|v_k\|_\Gcal^2)
$$
\begin{lemma} This is a norm, equal to the sum of singular values
when the two RKHS are spaces of linear functions on an Euclidean
space.
\end{lemma}

The main attractiveness of the trace norm is its convexity~\cite{Boyd2001Rank,Srebro2005Maximum}. However, the trace norm
does not readily yield a representer theorem, and as shown in Section~\ref{sec:rep_trace}, it is more practical to penalize
the trace norm of the matrix of estimates.

\section{Representer theorems}
In this section we explicitly state and prove representer theorems in tensor product spaces when a functional is minimized with rank constraints. These theorems underlie the algorithms proposed in the next section.

In a collaborative filtering task, the data usually have a matrix form, i.e., many $\xb$'s (resp. $\yb$'s)
are identical. We let $\xb_1,\dots,\xb_{n_\Xcal}$ denote the $n_\Xcal$ distinct values of elements of $\Xcal$ in
the training data, and, respectively,
$\yb_1,\dots,\yb_{n_\Ycal}$ denote the $n_\Ycal$ distinct values of elements of $\Ycal$. We assume that we have observations
of only a subset of $\{1,\dots,n_\Xcal\} \times  \{1,\dots,n_\Ycal\}$. We thus denote
$i(u)$ and $j(u)$ the indices of the $u$-th observation and $z_u$ the observed target.

\subsection{Classical representer theorem in the tensor product RKHS}
 Given a loss function
$\ell:\RR\times\RR\rightarrow\RR$, for example the square loss
function $\ell(z,z')=(z-z')^2$, a first classical approach to
learn dependencies between the pair $\br{\xb,\yb}$ and the
variable $z$ is to consider it as a supervised learning problem
over the product space $\Xcal\times\Ycal$, and for example to
search for a function in the RKHS of the product kernel which
solves the following problem:
\begin{equation}\label{eq:mintensor}
\min_{f\in\Ho} \cbr{\ovr{n}\Sum{u}{n} \ell \br{f\br{\xb_{i(u)},\yb_{j(u)}},z_{u}} + \lambda\nm{f}_{\otimes}^2}\;.
\end{equation}
By the representer theorem \cite{Kimeldorf1971Some} the solution of
(\ref{eq:mintensor}) has an expansion of the form:
$$
f\br{\xb,\yb} = \Sum{u}{n}\alpha_{u}k_{\otimes}\br{\br{\xb_{i(u)},\yb_{i(u)}},\br{\xb,\yb}} =\Sum{u}{n}\alpha_{u} k\br{\xb_{i(u)},\xb}g\br{\yb_{j(u)},\yb}\;,
$$ for some vector $\br{\alpha_{1},\ldots,\alpha_{n}}\in\RR^n$. Note that the number of parameters $\alpha$ is
the number of observed values. For many loss functions the problem (\ref{eq:mintensor})
boils down to classical machine learning algorithms such as support
vector machines, kernel logistic regression or kernel ridge
regression, which can be solved by usual implementations with the product
kernel (\ref{eq:tensorkernel}).

\subsection{Representer theorem with rank constraint}
In order to take advantage of the possible representation of our predictor as a sum of
a small number of factors, we propose to consider the
following generalization of (\ref{eq:mintensor}):
\begin{equation}\label{eq:minrank}
\min_{f\in\Ho,\text{ rank}(f)\leqslant p} \cbr{\ovr{n}\Sum{u}{n} \ell\br{f\br{\xb_{i(u)},\yb_{j(u)}},z_{u}} + \lambda\nm{f}_{\otimes}^2}\;.
\end{equation}

As the following proposition shows, the solution of this
constrained minimization problem can also be reduced to a
finite-dimensional optimization problem (see a proof in the
technical annex):

\begin{proposition}\label{prop:representer}
The optimal solution of (\ref{eq:minrank}) can be written as
$ f = \sum_{i=1}^p u_i \otimes v_i$, where
\begin{equation}\label{eq:alphabeta}
u_{i}\br{\xb}=\Sum{l}{n_\Xcal}\alpha_{li}k\br{{\xb}_{l},\xb}\quad\text{ and }\quad v_{i}\br{\yb}=\Sum{l}{n_\Ycal}\beta_{li}g\br{{\yb}_{l},\yb},\quad i=1,\ldots,p,
\end{equation}
where $\alpha \in \rb^{n_\Xcal \times p } $ and $\beta \in \rb^{n_\Ycal \times p } $
 \end{proposition}
This proposition is a crucial contribution of this paper as it
allows to learn the function $f$, by learning the coefficients
$\alpha$ and $\beta$:  denoting by $\alpha_k$ the $k$-th column of
$\alpha$ (similarly for $\beta$),  we obtain from Proposition
\ref{prop:representer} that an equivalent formulation of
(\ref{eq:minrank}) is:
\begin{equation}\label{eq:minrank2}
\min_{ \alpha \in \rb^{ n_\Xcal \times p  } , \beta \in \rb^{n_\Ycal \times p } }
\cbr{\ovr{n}\Sum{u}{n}
\ell\br{\Sum{k}{p} ( K \alpha_k )_{i(u)} ( G \beta_k)_{j(u)}, z_u } +
\lambda\Sum{i}{p}\Sum{j}{p} \alpha_i^\top K \alpha_j \beta_i^\top G \beta_j}
\end{equation}
where $K$ is the $n_\Xcal \times n_\Xcal$ kernel matrix for the elements of $\Xcal$ (similarly for $G$).
In order to link with the trace norm formulation in the next section, if we denote
$\gamma = \sum_{k=1}^p \alpha_k \beta_k^\top$,
we note that the $n_\Xcal \times n_\Ycal$ matrix of predicted values is equal
to $F = K \gamma G $
resulting in the following optimization problem:
\begin{equation}\label{eq:minrank3}
\min_{ \gamma = \sum_k \alpha_k \beta_k^\top }
\cbr{\ovr{n}\Sum{u}{n}
\ell\br{ ( K \gamma G  )_{i(u),j(u)} , z_u }}  +
\lambda \tr \gamma^\top K \gamma G\;.
\end{equation}

\subsection{Representer theorems and trace norm}
\label{sec:rep_trace}

The trace norm does not readily lead to a representer theorem and a finite dimensional optimization
problem. It is thus preferable to
 penalize the trace norm of the predicted values $F_{ij} = f(\xb_i,\yb_j)$, and minimize
\begin{equation}\label{eq:mintracenorm}
\min_{f\in\Ho} \cbr{\ovr{n}\Sum{u}{n} \ell\br{f\br{\xb_{i(u)},\yb_{j(u)}},z_{u}} + \mu \| (f(\xb_i,\yb_j))\|_\ast+  \lambda\nm{f}_{\otimes}^2}\;.
\end{equation}

We have the following representer theorem (whose proof is postponed to the technical annex) for the problem (\ref{eq:mintracenorm}):
\begin{proposition}
The optimal solution of (\ref{eq:mintracenorm}) can be written in the form
$f(x,y) = \sum_{i=1}^{n_\Xcal} \sum_{j=1}^{n_\Ycal}  \gamma_{ij} k(x,x_i) k(y,y_j)$, where $\gamma \in \rb^{n_\Xcal \times n_\Ycal}$.
\end{proposition}
The optimization problem can thus be rewritten as:

\begin{equation}\label{eq:minrank4}
\min_{ \gamma \in \rb^{n_\Xcal \times n_\Ycal}  }
\cbr{\ovr{n}\Sum{u}{n}
\ell\br{ ( K\gamma G  )_{i(u),j(u)} , z_u }  + \mu \| K \gamma G  \|_\ast +
\lambda \tr \gamma^\top K \gamma G}
\end{equation}

Note that in contrast to the finite representation without any
constraint on the rank or the trace norm (where the number of
parameters is the number of observed values), the number of
parameters is the total number of elements in the matrix,
and this method, though convex, is thus of higher computational complexity.

\subsection{Reformulation in terms of Kronecker products}
\paragraph{Kronecker products}
Given a matrix $B \in \rb^{m\times n}$ and a matrix $C \in \rb^{p
\times q}$, the Kronecker product $A= B \otimes C$ is a matrix in
$\rb^{ mp \times nq}$ defined by blocks of size $p \times q$ where
the block $(i,j)$ is $b_{ij} C$.

The most important properties are the following (where it is
assumed that all matrix operations are well-defined): $ (A \otimes
B) ( C \otimes D) =  AC \otimes BD$, $(A \otimes B)^{-1} = A^{-1}
\otimes B^{-1}$, $(A \otimes B)^{\top} = A^{\top} \otimes B^{\top}$,
and if $Y = CXB^\top \Leftrightarrow \text{vec}(Y) = ( B \otimes
C)  \text{vec}(X)$ where $\text{vec}(X)$ is the stack of columns
of $X$.

\paragraph{Reformulation}
We have $M = K \gamma G$, and thus $\text{vec}(M)$ is the vector
of predicted values for all pairs $(x_i,y_j)$ and is equal to
$\text{vec}(M) = (K \otimes G) \text{vec}(\gamma)$. The matrix $K
\otimes G$ is the kernel matrix associated with all pairs, for the
kernel $k_\otimes$. The results in this section does not provide
additional representational power beyond the kernel $k_\otimes$,
but present additional regularization frameworks for tensor
product spaces. In the next section, we tackle the
representational part of our contribution and show how the kernels
$k$ and $g$ can be tailored to the task of matrix completion.

\section{Kernels for matrix completion with attributes}
The three formulations (\ref{eq:mintensor}), (\ref{eq:minrank})
and (\ref{eq:mintracenorm}) differ in the way they handle
regularization. They all require a choice of kernel over $\Xcal$
and $\Ycal$ to define the RKHS norm in $\Ho$, which we discuss in
this section. The main theme of this section is the distinction
between kernels linked to the attributes and the kernel linked to
the identities of each different $\xb$ and $\yb$ (referred to as
Dirac kernels).

\paragraph{Dirac kernels}
In the standard collaborative filtering framework, where no
attribute over $\xb$ or $\yb$ is available, a natural kernel over
$\Xcal$ and $\Ycal$ is the Dirac kernel ($k_{Dirac}(\xb,\xb')=1$ if $\xb=\xb'$, $0$ otherwise).
If both $k$ and $g$ are Dirac kernels, then $k_{\otimes}$ is also
a Dirac kernel by (\ref{eq:tensorkernel}) and the classical
approach (\ref{eq:mintensor}) is irrelevant in that case (the
function $f$ being equal to $0$ on unseen examples). The low-rank
constraint added in (\ref{eq:minrank}) results in a relevant
problem: in fact for $\lambda=0^+$ we exactly recover the classical
low-rank matrix factorization problem.

\paragraph{Attribute kernels}
When attributes are available, they can be used to define kernels
which we denote $k_{Attributes}$ below. When both $k$ and $g$ are
kernels derived from attributes, then (\ref{eq:mintensor})
boils down to classical regression or classification over pairs.
Problem (\ref{eq:minrank}) provides an alternative problem,
where the rank of the function is constrained.

\paragraph{Multi-task learning}
Suppose now that attributes are available only for objects in
$\Xcal$, and not for $\Ycal$. It is then possible to take
$k=k_{Attribute}$ and $g=k_{Dirac}$. In that case the optimization
problem (\ref{eq:mintensor}) boils down to solving a classical
classification or regression problem for each value $y_{i}$
independently. Adding the rank constraint in (\ref{eq:minrank})
removes the independence among tasks by enforcing a decomposition
of the tasks into a limited number of factors, which leads to an
algorithm for multitask learning, based on a low-rank
representation of the predictor function. This approach is to be
contrasted with the framework of~\cite{Evgeniou2005Learning},
which is equivalent to $\Xcal$ finite of cardinality $p$, and
$k\br{\xb,\xb'}= 1-\lambda + \lambda p k_{Dirac}$. Our framework
focuses on a low rank representation for the predictor of each
task, while the framework of~\cite{Evgeniou2005Learning} focuses
on a set of predictor for each task that has small variance.
An extension of this multi-task learning framework, leading
to similar penalizations by trace norms, was independently derived by Argyriou et al.
\cite{massi}.

\paragraph{General formulation}
Supposing now that attributes are available on both $\Xcal$ and
$\Ycal$, let us consider the following interpolated kernels:
$$
\begin{cases}
k = \eta k_{Attribute}^x + (1-\eta) k_{Dirac}^x,\\
g = \zeta k_{Attribute}^y + (1-\zeta) k_{Dirac}^y,\\
\end{cases}
$$
where $0\leq\eta\leq 1$ and $0\leq\zeta\leq 1$. The resulting
product kernel is a sum of four terms:
$$
k_{\otimes} = \eta\zeta k_{Attribute}^x k_{Attribute}^y +
\eta(1-\zeta)k_{Attribute}^x k_{Dirac}^y +
$$
$$
+ (1-\eta)\zeta
k_{Dirac}^x  k_{Attribute}^y + (1-\eta)(1-\zeta) k_{Dirac}^x
k_{Dirac}^y .
$$
By varying $\eta$ and $\zeta$ this kernel provides an
interpolation between collaborative filtering ($\eta=\zeta=0$),
classical attribute-based regression on pairs ($\eta=\zeta=1$), and multi-task learning
($\eta=0$ and $\zeta=1$, or $\eta=1$ and $\zeta=0$).

In terms of kernel matrices on the set of all pairs, if we denote $K_{Att}$
the kernel matrix associated with the attributes associated with $\mathcal{X}$, and
$G_{Att}$
the kernel matrix associated with the attributes associated with $\mathcal{Y}$,
this is equivalent to using the matrix
$(
\eta K_{Att} + (1-\eta) \idm ) \otimes ( \zeta G_{Att}  + (1-\zeta) \idm ) =
\eta \zeta K_{Att} \otimes G_{Att} + \eta (1-\zeta) K_{Att} \otimes \idm  + (1-\eta)
\zeta \idm \otimes G_{Att} + (1-\eta)(1-\zeta) \idm \otimes \idm$,
which is the sum of four positive kernel matrices. The first one is simply the kernel
matrix for the tensor product space, while the last one is proportional to identity and usually appears
in kernel methods as the numerical effect of regularization~\cite{Cristianini2004}. The two
additional matrices makes the learning across rows and columns possible.

\paragraph{Generalization to new points}
One the usual drawbacks of collaborative filtering is the
impossibility to generalize to unseen data points (i.e., a new
movie or a new person in the context of movie recommendation).
When attributes are used, a prediction based on those can be made,
and thus using attributes has an added benefit beyond better
performance on matrix completion tasks.

\section{Algorithms}
In this section, we describe the algorithms used for the
optimization formulation in (\ref{eq:minrank2}) and
(\ref{eq:minrank4}). We also show that recent developments in
multiple kernel learning can be applied to both setting (enforced
rank constraint or trace norm).

\subsection{Fixed rank}
The function $(\alpha,\beta) \mapsto \ovr{n}\Sum{u}{n}
\ell\br{\Sum{k}{p} ( K \alpha_k )_{i(u)} ( L \beta_k)_{j(u)}, z_u
} + \lambda\Sum{i}{p}\Sum{j}{p} \alpha_i^\top K \alpha_j
\beta_i^\top G \beta_j$ is convex in each argument separately but
is not jointly convex. There are thus two natural optimization
algorithms: (1) alternate convex minimization with respect to
$\alpha$ and $\beta$, and (2) direct joint minimization using
Quasi-Newton iterative methods~\cite{bfgs}(in simulations we have
used the latter scheme).

As in \cite{Srebro2005Fast}, the fixed rank formulation, although
not convex, has the advantage of being parameterized by low-rank
matrices and is thus of much lower complexity than the convex
formulation that we know present. We present experimental results
in the next section using only this fixed rank formulation.

\subsection{Convex formulation}
If the loss $\ell$ is convex, then the function $\gamma \mapsto
\ovr{n}\Sum{u}{n} \ell\br{ ( K\gamma G  )_{i(u),j(u)} , z_u }  +
\mu \| K \gamma G  \|_\ast + \lambda \tr \gamma^\top K \gamma G $
is a convex function. However, even when the loss is
differentiable, the trace norm is non differentiable, which makes
iterative descent methods such as Newton-Raphson non
applicable~\cite{Boyd2003Convex}.

For specific losses which are SDP-representable, i.e., which can
represented by a semi-definite program, such as the square loss or
the hinge loss, the minimization of this function can be cast as
semi-definite program (SDP)~\cite{Boyd2003Convex}. For
differentiable losses which are not SDP-representable, such as the
logistic loss, an efficient algorithm is to modify the trace norm
to make it differentiable (see e.g \cite{Srebro2005Fast}). For
example, instead of penalizing the sum of singular values
$\lambda_i$, one may penalize the sum of $\sqrt{\lambda_i +
\varepsilon^2}$, which leads to a twice differentiable
function~\cite{lewis02twice}.

\subsection{Learning the kernels}
In this section, we show that the rank constraint and the trace
norm constraint can also be used in the multiple kernel learning
framework~\cite{BachLanc2004skm,Pontil}. We can indeed show that
if the loss $\ell$ is convex, then, as a function of the kernel
matrices, the optimal values of the optimization problems
(\ref{eq:minrank2}) and (\ref{eq:minrank4}) are convex functions,
and thus the kernel can be learned efficiently by minimizing those
functions. We do not use this method in our experiments, however,
we only include this for completeness.

\begin{proposition} Given $\beta_1,\dots,\beta_p$, the following function is convex in $K$:
$$
K \mapsto \min_{\alpha }
\cbr{\ovr{n}\Sum{u}{n}
\ell\br{\Sum{k}{p} ( K \alpha_k )_{i(u)} ( L \beta_k)_{j(u)}, z_u } +
\lambda\Sum{i}{p}\Sum{j}{p} \alpha_i^\top K \alpha_j \beta_i^\top G \beta_j}\;.
$$
The following function only depends on the Kronecker product $K \otimes G$ and is a convex function of $K \otimes G$.
$$
(K,G) \mapsto \min_{ \gamma \in \rb^{n_\Xcal \times n_\Ycal}  }
\cbr{\ovr{n}\Sum{u}{n}
\ell\br{ ( K\gamma G  )_{i(u),j(u)} , z_u }  + \mu \| K \gamma G  \|_\ast +
\lambda \tr \gamma^\top K \gamma G}\;.
$$
\end{proposition}
This proposition (whose proof is in the technical annex) shows that in the case of the rank constraint (\ref{eq:minrank2}), if we parameterize $K$ as $K=\sum_j
\eta_j K_j$, the weights $\eta$ and $\alpha$ can be learned
simultaneously~\cite{BachLanc2004skm,Pontil}. In particular, the
optimal weighting between attributes and the Dirac Kernel can be
learned directly from data. Note that a similar proposition holds
when the role of $x$ and $y$ are exchanged; when alternate
minimization is used to minimize the objective function, the
kernels can be learned at each step. In the case of the trace norm constraint (\ref{eq:minrank4}) it shows that we can learn a linear combination of basis
kernels, either the 4 kernels presented earlier obtained from Dirac's
and attributes, or more general combinations.  We leave this avenue open for future research.

\section{Experiments}

We tested the method on the well-known MovieLens 100k dataset from
the GroupLens Research Group at the University of Minnesota. This
dataset consists of ratings of 1682 movies by 943 users. The
ratings consisted of a score from the range 1 to 5, where 5 is the
highest ranking. Each user rated some subset of the movies, with a
minimum of 20 ratings per user, and the total number of ratings
available is exactly 100,000, averaging about 105 per user. To
speed up the computation, we used a random subsample of 800 movies
and 400 users, for a total of 20541 ratings. We divided this set
into 18606 training ratings and 1935 for testing. This dataset was
rather appropriate as it included attribute information for both
the movies and the users. Each movie was labelled with at least
one among $19$ genres (e.g., action or adventure), while users'
attributes age, gender, and an occupation among a list of $21$
occupations (e.g., administrator or artist).

We performed experiments using the rank constraint described in
\ref{eq:minrank}, and we used the more standard approach of cross
validation to choose kernel parameters. Thus, our method requires
selection of four parameters: the rank $d$ of the estimation
matrix; the regularization parameter $\lambda$; and the values
$\eta,\zeta \in [0,1]$, the tradeoff between the Dirac kernel and
Attribute kernel, for the users and movies respectively. The
parameters $\lambda$ and $d$ both act as regularization parameters
and we choose them using cross validation. The values $\eta,
\zeta$ were chosen out of $\{0, 0.15, 0.5, 0.85, 1\}$, the rank
$d$ ranged over $\{50, 80, 130, 200\}$, and $\lambda$ was chosen
from $\{25, 5, 1, 0.2, 0.04\} \times 10^{-6}$.

In Table 1, we show the performance for various choices or rank
$d$ and for various values of $\eta$ and $\zeta$, after selecting
$\lambda$ in each case using cross-validation. We also show in
bold the performance for the parameters selected using
cross-validation. Notice that, performance is consistently worse
when $\eta$ and $\zeta$ are chosen at the corners when compared
with values at the interior of $[0,1] \times [0,1]$. We observed
this to be true, in fact, not only when $d$ and $\lambda$ are
chosen by cross-validation, but for {\em every} choice of $d$ and
$\lambda$. Figure 1 shows the test mean squared error for $d$ and
$\lambda$ selected at each point using cross validation (Left), as
well as (Right) that for a fixed $d$ and $\lambda$ (the plot looks
similar for any fixed values of $d$ and $\lambda$) over the range
of $\eta$ and $\zeta$. Observe that the performance is best in the
interior of the $\eta$, $\zeta$ area, and worsens as we get
towards the edges and particularly at the corners. This is what we
might expect: at the corners we are no longer taking advantage of
either attribute information or ID information, either for the
class of movies or of the class of users.

Also notice in Table 1 that, as expected, regularization through
controlling the rank $d$ is indeed important. Regularization
through parameter $\lambda$ is also necessary: for $(d, \eta,
\zeta) = (130, 0.15, 0.15)$ shown in Table 1, test performance is
1.0351 when $\lambda = 0.2$, but is 1.1401 when $\lambda = 0.04$,
and 1.1457 when $\lambda = 1$ (we observe such changes in
performance across values of $\lambda$ for all other choices of
$d$, $\eta$, and $\zeta$). Hence, it is important to balance both
regularization terms. In fact, we use cross validation to select
all parameters.

\begin{table}[h]\label{tab:mse}
\begin{center}
\begin{tabular}{c|cccccc}
            &  $(\eta,\zeta) = (0,0)$
            &   $(0,1)$
            & $(1,0)$
            &  $(1,1)$
            &  $(0.5,0.5)$
            & $(0.15,0.15)$ \\ \hline

  $d = 50$  & 1.5391  & 1.6436  & 1.1999  & 1.1310 & 1.1106 & 1.0676 \\
  $d = 80$  & 1.5552  & 1.4008  & 1.2221  & 1.1138 & 1.0544 & 1.0478 \\
  $d = 130$ & 1.3294  & 1.3787  & 1.2315  & 1.0999 & 1.0611 & {\bf 1.0351} \\
  $d = 200$ & 1.3806  & 1.4234  & 1.2192  & 1.0818 & 1.0587 & 1.0596
\end{tabular}
\end{center}
\caption{Mean Squared Test Error results for various values of $\eta$ and
  $\zeta$ for three choices of rank $d$. In each of these, $\lambda$
  was chosen using cross-validation. Bold indicates the performance for the final parameters selected. }
\end{table}

\begin{figure}[h]\label{fig:etazeta}
%\begin{center}
\centerline{\psfig{file=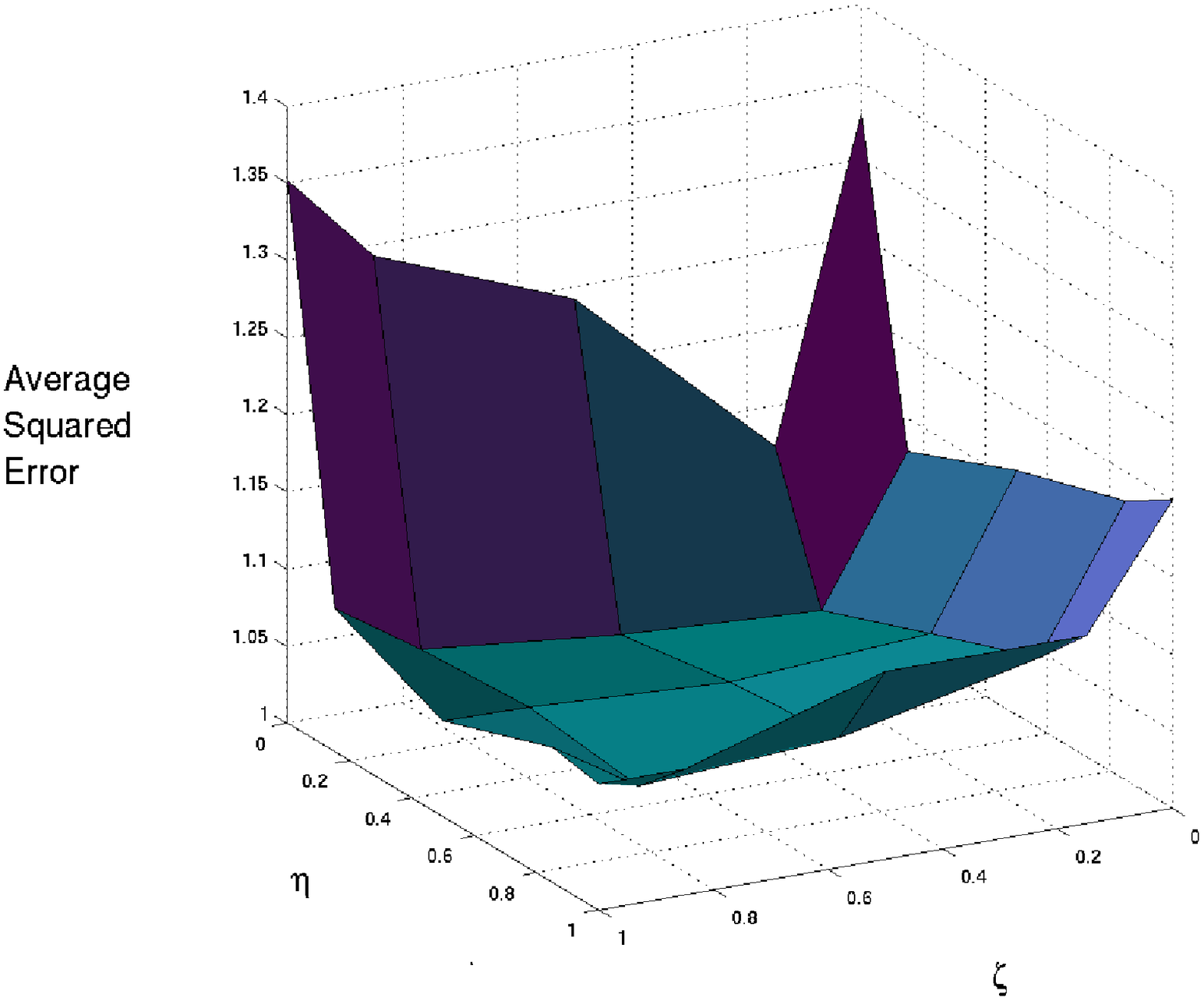,width=.49\textwidth}
\psfig{file=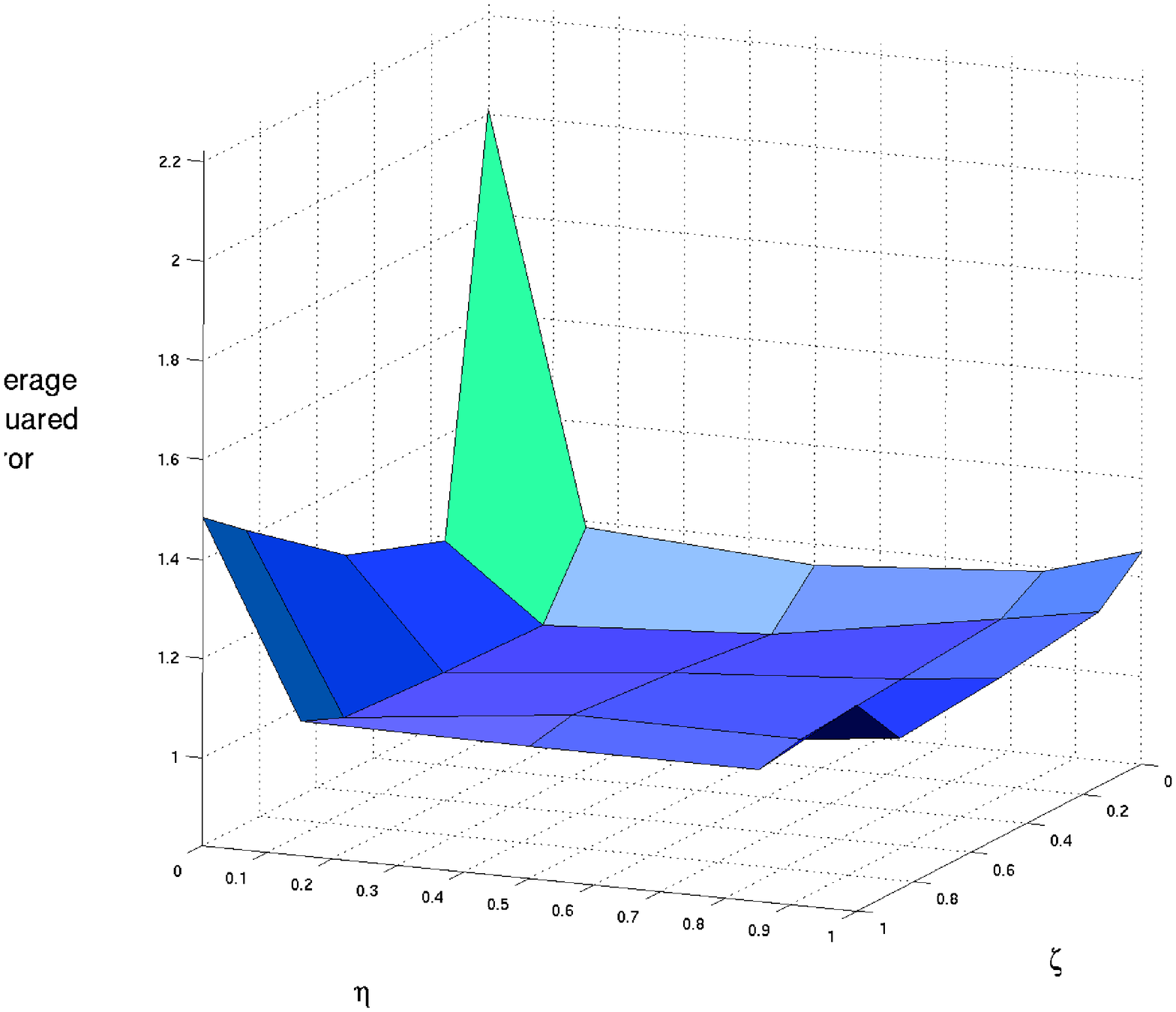,width=.49\textwidth}}
%\includegraphics[height=3in]{../../plots/color_eta_v_zeta.eps}
%\end{center}
\caption{{\it Left}: A plot of Mean Squared Error as we vary $\eta$ and
$\zeta$. We chose $d$ and $\lambda$ using cross-validation. {\it Right}:
    A plot of Mean Squared Error as we vary $\eta$ and
  $\zeta$ (for a fixed choice of $\lambda$ and $d$). In both cases we see
  the performance worsen at the extreme values when either $\eta$ or
  $\zeta$ become 0.}
\end{figure}

 \section{Conclusion}

We presented a method for solving a generalized matrix completion
problem where we have attributes describing the matrix
dimensions. Various approaches, such as standard low rank matrix
completion, are special cases of our method, and preliminary experiments confirm the benefits of our method.
%We note that the problem of matrix completion with attributes has recently deserved attention for the problem of gene network inference in computational biology \cite{Vert2005Supervised}, for the possibility to learn for multiple kernels is important.
An interesting direction of future research is to explore further the multi-task learning algorithm we obtained with low-rank constraint. On the
theoretical side, a better understanding of the effects of norm and rank regularizations and their interaction would be helpful.
% (e.g., \cite{ZwaldKPM}).
%\subsubsection*{Acknowledgments}
\small
\bibliographystyle{plain}
\bibliography{lowrank_techreport}

\appendix

\section{Rank of a function in a tensor product RKHS (Section 2.2)}

An element of $\Ho$ can always be decomposed as a possibly infinite sum of
atomic terms of the form $u(x)v(y)$ where $u \in \Kcal$ and $v \in
\Gcal$. We define the \emph{rank} $\text{rank}(f) \in \mathbb{N} \cup
\{\infty\}$ of an element $f$ of $\Ho$ as the minimal number of
atomic terms in any decomposition of $f$, i.e,
$\text{rank}(f)$ is the smallest integer $p$ such that $f$ can be
expanded as:
$$
f\br{\xb,\yb} = \Sum{1}{p}u_{i}\br{\xb}v_{i}\br{\yb}\;,
$$ for some functions $u_{1},\ldots,u_{p} \in\Kcal$ and
$v_{1},\ldots,v_{p}\in\Gcal$, if such an integer $p$ does exist (otherwise, the rank is infinite).

When the two RKHS are  spaces of linear functions on an Euclidean
space, then the tensor product can be identified to the space of bilinear forms on the product of the two Euclidean spaces, and the notion of rank coincides with the usual notion of rank for matrices. We note that an alternative characterization of $\text{rank}(f)$ is the
supremum of the ranks of the matrices
$M_{i,j}=f\br{\xb_{i},\yb_{j}}$ over the choices of finite sets
$\xb_{1},\ldots,\xb_{m}\in\Xcal$ and
$\yb_{1},\ldots,\yb_{p}\in\Ycal$ (see technical annex for a proof).

\begin{proposition}
$\text{rank}(f)$ is equal to the
supremum of the ranks of the matrices
$M_{i,j}=f\br{\xb_{i},\yb_{j}}$ over the choices of finite sets
$\xb_{1},\ldots,\xb_{m}\in\Xcal$ and
$\yb_{1},\ldots,\yb_{p}\in\Ycal$.
\end{proposition}
\begin{proof}
If $f$ can be expanded as $\sum_{i=1}^p u_{i}\otimes v_{i}$, then
for any choice of finite sets $\xb_{1},\ldots,\xb_{m}\in\Xcal$ and
$\yb_{1},\ldots,\yb_{p}\in\Ycal$, the matrix
$M_{i,j}=f\br{\xb_{i},\yb_{j}}$ can be expanded as $M=\sum_{i=1}^p
t_{i}w_{i}^\top$, where $t_{i,j}=u_{i}(x_{j})$ and
$w_{i,j}=v_{i}(y_{j})$, and has therefore a rank not larger than
$p$. Taking the smallest possible $p$, this shows that the rank of
$M$ can not be larger than the rank of $f$. Conversely, if
$f=\sum_{i=1}^p u_{i}\otimes v_{i}$ with $\text{rank}(f)=p$, we
need to show that there exist two sets of points such that the
corresponding matrix $M$ has rank at least $p$. We first observe
that both sets $\br{u_{i}}_{i=1,\ldots,p}$ and
$\br{v_{i}}_{i=1,\ldots,p}$ form linearly independent families in
$\Kcal$ and $\Gcal$, respectively. Indeed, if for example $u_{p} =
\sum_{i=1}^{p-1} r_{i}u_{i}$, then $f(x,y)=\sum_{i=1}^{p-1}
u_{i}(x)\sqb{v_{i}+r_{i}v_{p}}(y)$ which contradicts the
hypothesis that $\text{rank}(f)=p$. It is therefore possible to
find to sets of points $\xb_{1},\ldots,\xb_{p}\in\Xcal$ and
$\yb_{1},\ldots,\yb_{p}\in\Ycal$, such that the both sets of
vectors $\br{t_{i}}_{i=1,\ldots,p}$ and
$\br{w_{i}}_{i=1,\ldots,p}$ form linearly independent families in
$\RR^p$ (where $t_{i,j}=u_{i}(x_{j})$ and $w_{i,j}=v_{i}(y_{j})$).
The matrix $M$ corresponding to these sets of points has rank $p$.
\end{proof}

\section{Representer theorem with rank constraint (Proof of Proposition 1)}
Here we show that the solution of:
\begin{equation}\label{eq:app-minrank}
\min_{f\in\Ho,\text{ rank}(f)\leqslant p} \cbr{\ovr{n}\Sum{u}{n} \ell\br{f\br{\xb_{i(u)},\yb_{j(u)}},z_{u}} + \lambda\nm{f}_{\otimes}^2}\;,
\end{equation}
can be reduced to a finite-dimensional optimization problem:

\begin{proposition}\label{prop:app-representer}
The optimal solution of (\ref{eq:app-minrank}) can be written as
$ f = \sum_{i=1}^p u_i \otimes v_i$, where
\begin{equation}\label{eq:app-alphabeta}
u_{i}\br{\xb}=\Sum{l}{n_\Xcal}\alpha_{li}k\br{{\xb}_{l},\xb}\quad\text{ and }\quad v_{i}\br{\yb}=\Sum{l}{n_\Ycal}\beta_{li}g\br{{\yb}_{l},\yb},\quad i=1,\ldots,p,
\end{equation}
where $\alpha \in \rb^{n_\Xcal \times p } $ and $\beta \in \rb^{n_\Ycal \times p } $
 \end{proposition}
\begin{proof}
Let $\Ho^S$ denote the subspace of $\Ho$ spanned by the functions
$k_{\otimes}\br{\br{x_{i},y_{j}},.}$ for $i=1,\ldots,n_{\Xcal}$
and $j=1,\ldots,n_{\Ycal}$, and $\Ho^\perp$ denote its orthogonal
supplement ($\Ho=\Ho^S\oplus\Ho^\perp$). Similarly, let $\Kcal^S$
and $\Gcal^S$ denote respectively the subspace of $\Kcal$ and
$\Gcal$ spanned by the functions $k\br{\xb_{i},.}$ for
$i=1,\ldots,n_{\Xcal}$, (resp. $g\br{\yb_{i},.}$ for
$i=1,\ldots,n_{\Ycal}$) and $\Kcal^\perp$ and $\Gcal^\perp$  the
corresponding orthogonal supplements in $\Kcal$ and $\Gcal$. Any
function $f\in\Ho$ of rank less than $p$ can be expanded as
$f=\Sum{i}{p}u_{i}\otimes v_{i}$, with $u_{i}\in\Kcal$ and
$v_{i}\in\Gcal$. Now, denote by $u_{i}=u_{i}^S + u_{i}^\perp$ the
unique decomposition of $u_{i}$ over
$\Kcal=\Kcal^S\oplus\Kcal^\perp$, and define similarly
$v_{i}=v_{i}^S + v_{i}^\perp$ with $v_{i}^S\in\Gcal^S$ and
$v_{i}^\perp\in\Gcal^\perp$. We therefore obtain:
\begin{equation}\label{eq:app-decomposition}
f=\Sum{i}{p}u_{i}^{S}\otimes v_{i}^{S} + \Sum{i}{p}u_{i}^{S}\otimes v_{i}^{\perp} + \Sum{i}{p}u_{i}^{\perp}\otimes v_{i}^{S} + \Sum{i}{p}u_{i}^{\perp}\otimes v_{i}^{\perp}.
\end{equation}
We now claim that the last three terms in (\ref{eq:app-decomposition})
are in $\Ho^\perp$. Indeed, taking for example a term
$u_{i}^{S}\otimes v_{i}^\perp$, we have $u_{i}^{S}\otimes
v_{i}^\perp\br{\xb_{j},\yb_{l}} = u_{i}^{S}\br{\xb_{j}}
v_{i}^\perp\br{\yb_{l}} = 0$ because $v_{i}^\perp\br{\yb_{i}}=0$,
and therefore $u_{i}^{S}\otimes v_{i}^\perp \in \Ho^\perp$. A
similar computation shows that $u_{i}^{\perp}\otimes v_{i}^{S}$
and $u_{i}^{\perp}\otimes v_{i}^{\perp}$ are both in $\Ho^\perp$.
On the other hand, because $u_{i}^S \in\Kcal^S$ and
$v_{i}^S\in\Gcal^S$, one easily gets that $u_{i}^S\otimes v_{i}^S
\in\Ho^S$. Therefore $\Sum{i}{p}u_{i}^S v_{i}^S$ is the orthogonal
projection of $s$ onto $\Ho^S$ and is of rank at most $p$. We can
conclude like for the classical representer theorem that is $f$ is
 not restricted to $\Sum{i}{p}u_{i}^S v_{i}^S$, then
$\Sum{i}{p}u_{i}^S v_{i}^S$ provides a rank $p$ function of $\Ho$
with a strictly smaller functional value, leading to a
contradiction.
\end{proof}

\section{Representer theorems and trace norm (proof of Proposition 2)}
\label{sec:app-rep_trace}

Here we show a representer theorem for the solution of:
\begin{equation}\label{eq:app-mintracenorm}
\min_{f\in\Ho} \cbr{\ovr{n}\Sum{u}{n} \ell\br{f\br{\xb_{i(u)},\yb_{j(u)}},z_{u}} + \mu \| (f(\xb_i,\yb_j))\|_\ast+  \lambda\nm{f}_{\otimes}^2}\;.
\end{equation}
\begin{proposition}
The optimal solution of (\ref{eq:app-mintracenorm}) can be written in the form
$f(x,y) = \sum_{i=1}^{n_\Xcal} \sum_{j=1}^{n_\Ycal}  \gamma_{ij} k(x,x_i) k(y,y_j)$, where $\gamma \in \rb^{n_\Xcal \times n_\Ycal}$.
\end{proposition}
\begin{proof}
Our objective function is the sum of a function of values of $f$ for all pairs $(x_i,y_j)$ plus a squared RKHS norm. The usual representer
in the RKHS associated with $k_\otimes$
then applies, and we get a solution of the form
$f(x,y) = \sum_{i=1}^{n_\Xcal} \sum_{j=1}^{n_\Ycal}  \gamma_{ij} k_\otimes( (x,y),(x_i,y_j))
= \sum_{i=1}^{n_\Xcal} \sum_{j=1}^{n_\Ycal}  \gamma_{ij} k(x,x_i) k(y,y_j)$ where  $\gamma \in \rb^{n_\Xcal \times n_\Ycal}$.
\end{proof}

\section{Learning the kernels (proof of Proposition 3)}

In this section we prove that
if the loss $\ell$ is convex, then, as a function of the kernel
matrices, the optimal values of the optimization problems proposed in the paper are convex functions,

\begin{proposition} Given $\beta_1,\dots,\beta_p$, the function
$$H: K \mapsto \min_{\alpha }
\cbr{\ovr{n}\Sum{u}{n}
\ell\br{\Sum{k}{p} ( K \alpha_k )_{i(u)} ( L \beta_k)_{j(u)}, z_u } +
\lambda\Sum{i}{p}\Sum{j}{p} \alpha_i^\top K \alpha_j \beta_i^\top G \beta_j}
$$
is convex in $K$.
\end{proposition}
\begin{proof} Given $\beta$, the objective function is
convex and thus (under appropriate classical conditions), the
minimum value is equal to the maximum value of the dual problem,
obtained by adding variables $q_u$ and adding constraints $q_u=
\Sum{k}{p} ( K \alpha_k )_{i(u)} ( L \beta_k)_{j(u)}$, together
with the appropriate Lagrange multipliers. The results follow from
derivations obtained in~\cite{BachLanc2004skm,Pontil}.
\end{proof}

\begin{proposition} The function
\begin{equation}\label{eq:app-minrank4}
H: (K,G) \mapsto \min_{ \gamma \in \rb^{n_\Xcal \times n_\Ycal}  }
\cbr{\ovr{n}\Sum{u}{n}
\ell\br{ ( K\gamma G  )_{i(u),j(u)} , z_u }  + \mu \| K \gamma G  \|_\ast +
\lambda \tr \gamma^\top K \gamma G}
\end{equation}
only depends on the Kronecker product $K \otimes G$ and is a convex function of $K \otimes G$.

\end{proposition}
\begin{proof}
The objective function (\ref{eq:app-minrank4}) was originally obtained
from (\ref{eq:app-mintracenorm}), which is the sum of a term that is a convex function of the values of
a function $f\in \Ho$, for all pairs $(\xb_i,\yb_j)$, and the norm
$\| f\|^2_\otimes$.
The results of~\cite{Pontil} applies to this
case and thus the minimum value is a convex function in the kernel
matrix for all pairs $(\xb_i,\yb_j)$ and the kernel $k_\otimes$,
which is exactly $K \otimes G$.
\end{proof}

\end{document}